\documentclass[runningheads]{llncs}
\usepackage{graphicx,wrapfig}
\usepackage{caption}
\usepackage{subcaption}
\usepackage{geometry}
\usepackage{amsmath}
\usepackage{amsfonts}
\usepackage{mathtools}
\usepackage{hyperref}   
\hypersetup{
  colorlinks   = true,
  urlcolor     = blue,
  linkcolor    = blue,
  citecolor    = red
}
\usepackage{adjustbox}
\usepackage[dvipsnames]{xcolor}
\usepackage{authblk}
\usepackage{siunitx}
\usepackage[subtle]{savetrees}
\usepackage{lipsum}
\usepackage{booktabs}
\usepackage{float}
\usepackage{booktabs}
\begin{document}

\makeatletter
\newcommand\blfootnote[1]{%
  \begingroup
  \renewcommand{\@makefntext}[1]{\noindent\makebox[1.8em][r]#1}
  \renewcommand\thefootnote{}\footnote{#1}%
  \addtocounter{footnote}{-1}%
  \endgroup
}
\makeatother

\title{Learning in Hybrid Active Inference Models}

\author{%
Poppy Collis \thanks{Corresponding author} \inst{1,\dag}  \and 
Ryan Singh \inst{1,2, \dag} \and 
Paul F Kinghorn \inst{1} \and
Christopher L Buckley \inst{1,2}
}%
\institute{
School of Engineering and Informatics, University of Sussex, Brighton, UK\\
\email{ \{pzc20, rs773, p.kinghorn, c.l.buckley\}@sussex.ac.uk},\\ 
\and
VERSES AI Research Lab, Los Angeles, California, USA\\}
\authorrunning{P. Collis, R. Singh, P. F. Kinghorn and C.L. Buckley}

\maketitle

\blfootnote{- $\dag$ Equal contribution }

\begin{abstract}
An open problem in artificial intelligence is how systems can flexibly learn discrete abstractions that are useful for solving inherently continuous problems. Previous work in computational neuroscience has considered this functional integration of discrete and continuous variables during decision-making under the formalism of active inference \cite{friston2017graphical,parr2018discrete}.
However, their focus is on the expressive physical implementation of categorical decisions and the hierarchical mixed generative model is assumed to be known. As a consequence, it is unclear how this framework might be extended to the learning of appropriate coarse-grained variables for a given task. In light of this, we present a novel hierarchical hybrid active inference agent in which a high-level discrete active inference planner sits above a low-level continuous active inference controller. We make use of recent work in recurrent switching linear dynamical systems (rSLDS) which learn meaningful discrete representations of complex continuous dynamics via piecewise linear decomposition \cite{lindermanRecurrentSwitchingLinear2016}. The representations learnt by the rSLDS inform the structure of the hybrid decision-making agent and allow us to (1) lift decision-making into the discrete domain enabling us to exploit information-theoretic exploration bonuses (2) specify temporally-abstracted sub-goals in a method reminiscent of the options framework \cite{SUTTON1999181} and (3) `cache' the approximate solutions to low-level problems in the discrete planner. We apply our model to the sparse Continuous Mountain Car task, demonstrating fast system identification via enhanced exploration and successful planning through the delineation of abstract sub-goals.

\end{abstract}

\section{Introduction}
In a world that is inherently high-dimensional and continuous, the brain’s capacity to distil and reason about discrete concepts represents a highly desirable feature in the design of autonomous systems. Humans are able to flexibly specify abstract sub-goals during planning, thereby reducing complex problems into manageable chunks \cite{newell1972human,Gobet2001}.  Indeed, translating problems into discrete space offers distinct advantages in decision-making systems. For one, discrete states admit the direct implementation of classical techniques from decision theory such as dynamic programming \cite{LaValle_2006}. Furthermore, we also find the computationally feasible application of information-theoretic measures (e.g. information-gain) in discrete spaces. Such measures (generally) require approximations in continuous settings but these have closed-form solutions in the discrete case \cite{fristonstructure}. While the prevailing method for translating continuous variables into discrete representations involves the simple grid-based discretisation of the state-space, this becomes extremely costly as the dimensionality increases \cite{coulom2007,Mnih2015HumanlevelCT}. We therefore seek to develop a framework which is able to smoothly handle the presence of continuous variables whilst maintaining the benefits of decision-making in the discrete domain.

\begin{figure}[h!]
    \includegraphics[width=10cm]{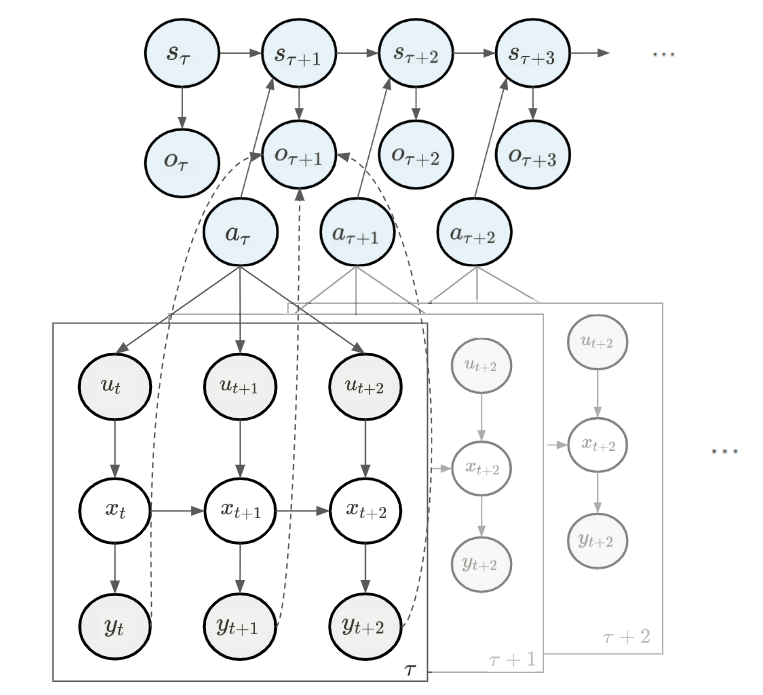}
    \centering
\caption{\textbf{Previous discrete-continuous active inference models have focused on the physical implementation of categorical decisions in continuous space}. Here, outcomes from the high-level active inference planner select from a set of discrete models of continuous dynamics, speciﬁed by a prior over their hidden causes. This mixed generative model effectively generates discrete sequences of short continuous trajectories deﬁned in terms of their generalised coordinates of motion. The discrete planner is formulated as a standard POMDP generative model (see Sec.~\ref{discrete-planner-sec}) with discrete states $s_\tau$ and observations $o_\tau$. The first action $a_\tau$ of the selected policy (see Sec.~\ref{aif-sec}) is then passed down to the continuous active inference controller via the expected observation $q(o_{\tau+1}|a_\tau)$. This distribution weights a set of fixed point means $\{\eta_m\}_{m = 0}^{M-1}$ which map the $m$ discrete latent states into continuous state space. The resulting weighted average, $\eta_\tau =  \sum_m \eta_m \cdot q(o_{\tau+1,m}|a)$, serves as the mean of a prior over hidden causes, $p(\nu) = \mathcal N(\eta_{\tau}, \pi_{\tau}^{-1})$, which drives the dynamics of the low-level continuous latent states $\tilde{x_t} = \{x_t, x_t^\prime, x_t^{\prime\prime}\}$ and observations $\tilde{y_t} = \{y_t, x_t^\prime, x_t^{\prime\prime}\}$ represented in generalised coordinates. Inherently, there is a separation of timescales in this open-loop control setup: the discrete controller sends an action down to the continuous controller which is then executed in a ballistic manner over several timesteps. After this low-level inner-loop completes, a process of Bayesian model selection is used to infer the current discrete state description of the low-level system given the trajectory of continuous observations $\tilde{y_t}$. This is then given as an observation for the discrete planner at the top. For a full treatment of this model, see \cite{friston2017graphical}.}
\label{discrete-cont}
\end{figure}

\subsection{Hybrid Active Inference}

Here, we draw on recent work in active inference (AIF) which has foregrounded the utility of decision-making in discrete state-spaces \cite{DACOSTA2020102447,fristonstructure}. Additionally, discrete AIF has been successfully combined with low-level continuous representations and used to model a range of complex behaviour including speech production, oculomotion and tool use \cite{friston2017graphical,parr2018discrete,friston2021active,parr2019computational,priorelli2024hierarchical}. As detailed in \cite{friston2017graphical}, such mixed generative models focus on the physical implementation of categorical decisions. This treatment begins with the premise that the world can be described by a set of discrete states evolving autonomously and driving the low-level continuous states by indexing a set of attractors (c.f. subgoals) encoded through priors which have been built into the model (see Fig.~\ref{discrete-cont}). While the emphasis of the above work is on mapping categorical decision-making to the continuous physical world, here, we approach the question of learning the generative model. Specifically, we seek the complete learning of appropriate discrete representations of the underlying dynamics and their manifestation in continuous space. Importantly, unlike the previous work mentioned here, we focus on instances in which the mapping between the discrete states and the continuous states is not assumed to be known. In this case, however, the assumption that higher-level discrete states autonomously drive lower-level continuous states (i.e. downward causation) becomes problematic. Any failure of the continuous system to carry out a discrete preference must be treated as an autonomous failure at the discrete level. Although useful for planning, this decoupling of the discrete from the continuous components makes it difficult to represent complex dynamics, which in turn creates difficulties in learning. 

\begin{figure}[t]
    \includegraphics[width=6cm]{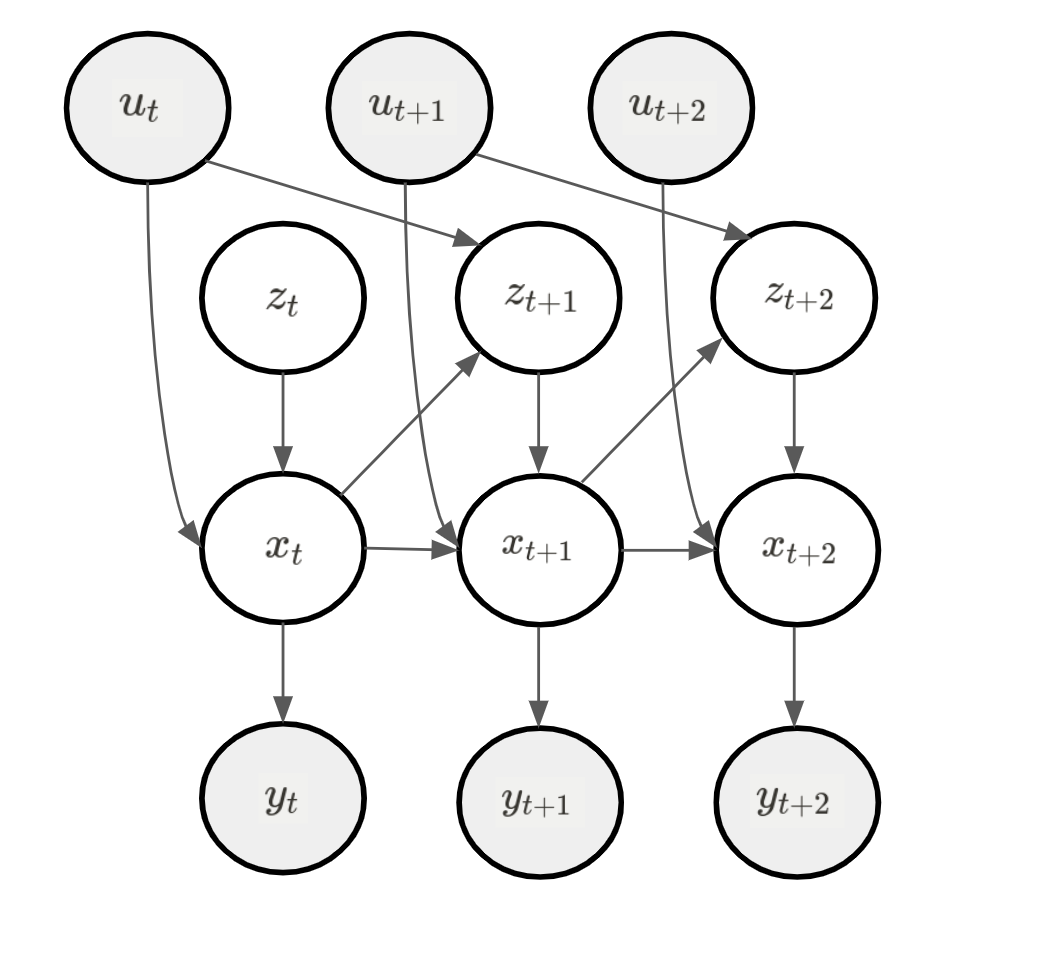}
    \centering
\caption{\textbf{Recurrent switching linear dynamical systems (rSLDS) discover meaningful discrete states and explain how their switching behaviour depends on continuous latent states}. This class of hybrid state space model includes a recurrent dependency of the discrete latent state $z_{t+1}$ on the continuous latent state $x_t$ and control input $u_t$. As in a standard SLDS, the continuous latent dynamics are conditionally linear (dependent on the current discrete state $z_t$) and generate observations $y_t$. Note that this figure shows the recurrent-only formulation of the rSLDS (see Sec.~\ref{rslds-section}) in which the discrete latent $z_t$ have no dependency on $z_{t-1}$ as is present in its canonical form.}

\label{rslds-dag}
\end{figure}

\subsection{Recurrent Switching Systems}

Previous work has demonstrated that models involving autonomous switching systems are often not sufficiently expressive to approximate realistic generative processes \cite{lindermanRecurrentSwitchingLinear2016}. They study this problem in the context of a class of hybrid state-space model known as switching linear dynamical systems (SLDS). These models have been shown to discover meaningful behavioural modes and their causal states via the piecewise linear decomposition of complex continuous dynamics \cite{hintonghahramani,NIPS2008_FOX}. The authors of \cite{lindermanRecurrentSwitchingLinear2016} remedy the problem associated with limited expressivity by introducing \textit{recurrent} switching linear dynamical systems (rSLDS) (see Fig.~\ref{rslds-dag}). These models importantly include a dependency from the underlying continuous variables in the high-level discrete transition probabilities. By providing an understanding of the continuous latent causes of switches between the discrete states via this additional dependency, the authors demonstrate improved generative capacity and predictive performance. We propose this richer representation can be useful for decision making and control. This recurrent transition structure can be exploited such that continuous priors can be flexibly specified for a low-level controller in order to drive the system into a desired region of the state space. Using statistical methods to fit these models not only liberates us from the need to explicitly specify a mapping between discrete and continuous states a priori, but enables effective online discovery of useful non-grid discretisations of the state-space. 

\subsection{Emergent descriptions for planning}
Unfortunately, the inclusion of recurrent dependencies also destroys the neat separation of discrete planning from continuous control, creating unique challenges in performing roll-outs. Our central insight is to re-instate the separation by lifting the dynamical system into the discrete domain \textit{only during planning}. We do this by approximately integrating out the continuous variables, naturally leading to spatio-temporally abstracted actions and sub-goals. Our discrete planner therefore operates purely at the level of a re-description of the discrete latents, modelling nothing of the autonomous transition probabilities but rather reflecting transitions that are possible given the discretisation of the continuous state-space. In short, we describe a novel hybrid hierarchical active inference agent \cite{parr2022active} in which a discrete Markov decision process (MDP), informed by the representations of an rSLDS, interfaces with a continuous active inference controller implementing closed-loop control. We demonstrate the efficacy of this algorithm by applying it to the classic control task of Continuous Mountain Car \cite{openai_continuous_mountain_car}. We show that the exploratory bonuses afforded by the emergent discrete piecewise description of the task-space facilitates fast system identification. Moreover, the learnt representations enable the agent to successfully solve this non-trivial planning problem by specifying a series of abstract subgoals.

\section{Related work}

Such temporal abstractions are the focus of Hierarchical reinforcement learning (HRL), where high-level controllers provide the means for reasoning beyond the clock-rate of the low-level controllers primitive actions. \cite{dayanhinton,SUTTON1999181,Daniel2016,hafner2022deep}. The majority of HRL methods, however, depend on domain expertise to construct tasks, often through manually predefined subgoals as seen in \cite{Tessler_2017}. Further, efforts to learn hierarchies directly in a sparse environment have typically been unsuccessful \cite{vezhnevets17a}. In contrast, our abstractions are a natural consequence of lifting the problem into the discrete domain and can be learnt independently of reward. In the context of control, hybrid models in the form of piecewise affine (PWA) systems have been rigorously examined and are widely applied in real-world scenarios \cite{sontag,Bemporad2000,Borrelli2006}. Previous work has applied a variant on rSLDS (recurrent autoregressive hidden Markov models) to the optimal control of general nonlinear systems \cite{10128705,pmlr-v120-abdulsamad20a}. The authors use these models to the approximate expert controllers in a closed-loop behavioural cloning context. While their algorithm focuses on value function approximation, in contrast, we learn online without expert data and focus on flexible discrete planning. 

\section{Framework}

The following sections detail the components of our Hierachical Hybrid Agent (HHA). For additional information, please refer to Appendix.~\ref{appendix}.

\subsection{Generative Model: rSLDS(ro)}
\label{rslds-section}
In the recurrent-only (ro) formulation of the rSLDS (see Fig.~\ref{rslds-dag}), the discrete latent states $z_t \in \{1, 2, . . . , K\}$ are generated as a function of the continuous latents $x_t \in \mathbb R^M$ and the control input $u_t \in \mathbb R^N$ (specified by some controller) via a softmax regression model,
\begin{align}
    P(z_{t+1}|x_t, u_t) = softmax(W_x x_t + W_u u_t + r)
\end{align}
whereby $W_x \in \mathbb R^{K \times M}$ and $W_u \in \mathbb R^{K \times N}$ are weight matrices and $r$ is a bias of size $\mathbb R^{K}$. The continuous latent states $x_t$ evolve according to a linear dynamical system indexed by the current discrete state $z_t$.
\begin{equation}
\begin{split}
    x_{t+1}|x_t, u_t, z_t &= A_{z_{t}} x_t + B_{z_{t}} u_t + b_{z_t} + \nu_t, \\
    &\nu_t \sim \mathcal{N}(0, Q_{z_{t}})
\end{split}
\end{equation}
\begin{equation}
    y_t| x_t = C_{z_{t}}x_t + \omega_t, \;\;\;\; \omega_t \sim \mathcal{N}(0, S_{z_{t}})
\end{equation}
$A_{z_t}$ is the state transition matrix, which defines how the state $x_t$ evolves in the absence of input. $B_{z_t}$ is the control matrix which defines how external inputs influence the state of the system while $b_{z_t}$ is an offset vector. At each time-step $t$, we observe an observation $y_t \in \mathbb R^M$ produced by a simple linear-Gaussian emission model with an identity matrix $C_{z_t}$. Both the dynamics of the continuous latents and the observations are perturbed by zero-mean Gaussian noise with covariance matrices of $Q_{z_t}$ and $S_{z_t}$ respectively.

Inference requires approximate methods given that the recurrent connections break conjugacy rendering the conditional likelihoods non-Gaussian. Therefore, a Laplace Variational Expectation Maximisation (EM) algorithm is used to approximate the posterior distribution over the latent variables by a mean-field factorisation into separate distributions for the discrete states $q(z)$ and the continuous states $q(x)$. The discrete state is updated via a coordinate ascent variational inference (CAVI) approach by leveraging the forward-backward algorithm. The continuous state distribution is updated using a Laplace approximation around the mode of the expected log joint probability. This involves finding the most likely continuous latent states by maximizing the expected log joint probability and computing the Hessian to approximate the posterior. Full details of the Laplace Variational EM used for learning are given in \cite{zoltowski2020unifying}. 

The rSLDS is initialised according to the procedure outlined in \cite{lindermanRecurrentSwitchingLinear2016}. In order to learn the rSLDS parameters using Bayesian updates, conjugate matrix normal inverse Wishart (MNIW) priors are placed on the parameters of the dynamical system and recurrence weights. We learn the parameters online via observing the behavioural trajectories of the agent and updating the parameters in batches (every 1000 timesteps of the environment).

\subsection{Active Inference}
\label{aif-sec}

Equipped with a generative model, active inference specifies how an agent can solve decision making tasks \cite{parr2022active}. Policy selection is formulated as a search procedure in which a free energy functional of predicted states is evaluated for each possible policy. Formally, we use an upper bound on the expected free energy ($\mathcal G$) given by:

\begin{equation}
\label{G_decomposed_fully}
\begin{aligned}
\mathcal{G}_{1:T}(\pi)\le & \underbrace{-\mathbb{E}_{Q(\bold o|\pi)}[D_{KL} \left[ Q(\bold s| \bold o,\pi)\parallel Q(\bold s|\pi)\right]]}_\text{State Information Gain} \\
& \underbrace{-\mathbb{E}_{Q(\bold o|\pi)}[D_{KL} \left[ Q(\theta|\bold o,\pi)\parallel Q(\theta|\pi)\right]]}_\text{Parameter Information Gain} \\
& \underbrace{-\mathbb{E}_{Q(\bold o|\pi)}[\ln \tilde{p}(o)]}_\text{Utility}.
\end{aligned}
\end{equation}
Where $\bold s = \{s_1,...,s_T\}$ and $\bold o = \{o_1,...,o_T\}$ are the states and observations being evaluated under a particular policy or sequence of actions, $\pi = a_{1:T}$. The integration of rewards in the inference procedure is achieved by biasing the agent's generative model with an optimistic prior  over observing desired outcomes $\tilde{p}(o)$. Action selection then involves converting this into a probability distribution over the set of policies and sampling from this distribution accordingly.

\begin{figure}[h!]
    \includegraphics[width=10cm]{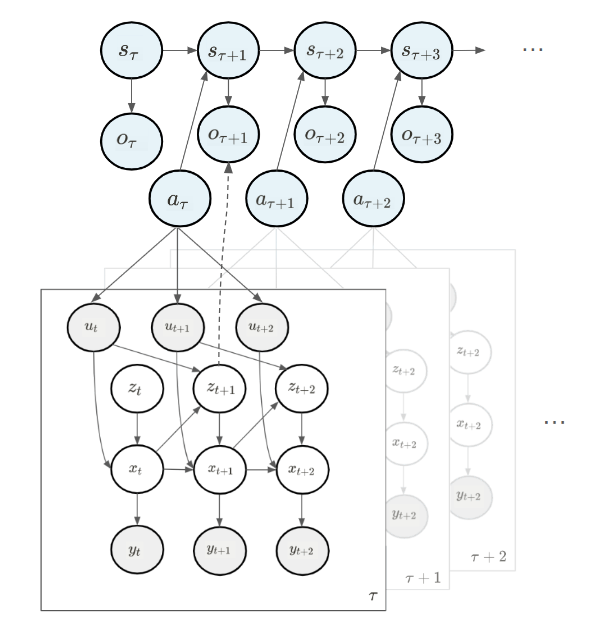}
    \centering
\caption{\textbf{Our Hybrid Hierarchical Agent learns emergent coarse-grained descriptions of the continuous state-space for planning and control}. Like previous work on mixed generative models in active inference shown in Fig.~\ref{discrete-cont}, we have a discrete active inference planner sitting above a low-level continuous active inference controller. The discrete planner is constructed as a standard POMDP generative model (see Sec. 3.3) with discrete states $s_\tau$ and observations $o_\tau$. However, our model departs from \cite{friston2017graphical} in the generation of coarse-grained variables which instead emerge from the the underlying rSLDS generative model. Here, the states $S$ of the planner are essentially a re-description of the discrete states $Z$ found by the rSLDS. The transition model probabilities are then constrained to reflect the adjacency structure of the polyhedral partitions of the state-space found by the softmax regression component of rSLDS. The chosen action $a_\tau$ from the high-level planner selects from a discrete set of continuous active inference controllers based on both the linear dynamics of the current discrete state $z_\tau$ and a control prior for the desired next discrete state. Using the rSLDS generative model, this prior is a flexibly specified continuous point in the state-space that is in the discrete region the agent wishes to move into (see Eq.~\ref{gen-ctl-prior-eq}). Unlike the models in \cite{friston2017graphical}, the action $a_\tau$ is temporally abstracted with no pre-defined timescale at the lower level. Instead, the discrete planner is only re-triggered when the system enters a new discrete state (i.e. $z_{t} \neq z_{t-1}$). At which point, the planner observes the new discrete state $z_\tau$ of the system and constructs a plan accordingly.}
\label{hha-diagram}
\end{figure}

\subsection{Discrete Planner}
\label{discrete-planner-sec}
\begin{figure}[b]
    \vskip 0.2in
    \begin{center}
        \centerline{\includegraphics[width=1.0\textwidth]{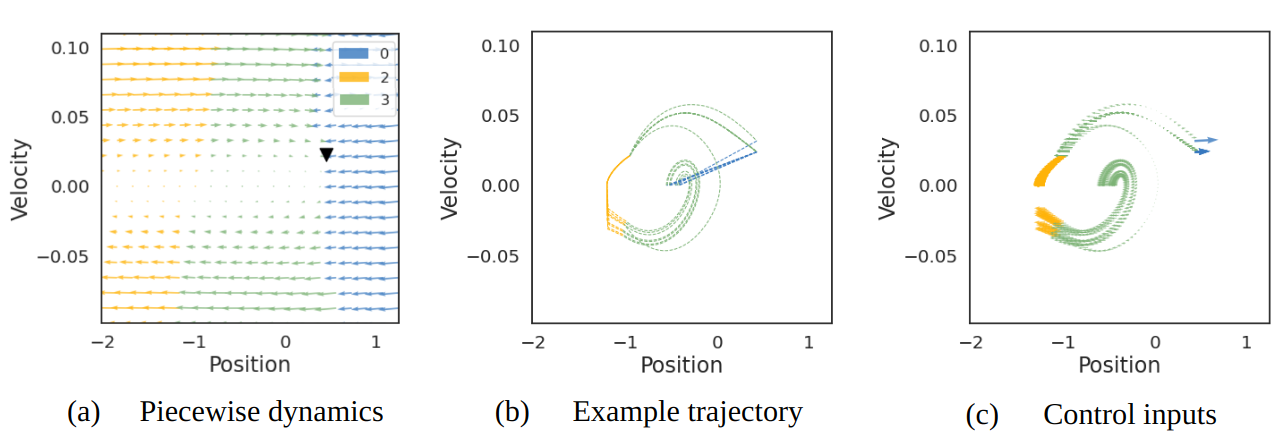}}
        \caption{\textbf{HHA solves nonlinear problems via specifying abstract sub-goals in state-space.} (a) Piecewise linear dynamics  of the Continuous Mountain Car state-space found by rSLDS represented as a vector plot where magnitude of the arrows indicates how fast the state is changing at that point. Reward location shown (\emph{black triangle}). While the rSLDS retrives 5 modes in total, here we plot only the modes seen in the position-velocity ($x$) space without showing the control input ($u$) axis. (b) Example trajectory (segments coloured by mode) showing the HHA consistently navigating to the goal. (c) Continuous control input (coloured by discrete action specified by planner and arrow size indicating magnitude and direction) over same example trajectory in (b).}
        \label{piecewise}
    \end{center}
    \vskip -0.2in
\end{figure}

In order to create approximate plans at the discrete level, we derive a high-level planner based on a re-description of the discrete latent states found by the rSLDS by approximately `integrating out' the continuous variables and the continuous prior. This process involves calculating the expected free energy ($\mathcal G$) for a continuous controller to drive the system from one mode to another. Importantly, the structure of the lifted discrete state transition model has been constrained by the polyhedral partition of the continuous state space extracted from the parameters of the rSLDS \footnote{For a visualisation of this partitioning of the state space, see Fig.~\ref{piecewise}(a)}: invalid transitions are assigned zero probability while valid transitions are assigned a high probability. In order to generate the possible transitions from the rSLDS, we calculate the set of active constraints for each region from the softmax representation, $p(z\mid x) = \sigma(Wx +  b)$. Specifically, to check that the region $i$ is adjacent to region $j$, we verify the solution using a linear program,
\begin{align}
    - b_j  = &\min (W_i - W_j ) x \\
    & \text{s.t. } (W_i - W_k)x \leq (b_i - b_k) \text{ }\forall k \in [K] \\
    & \text{s.t. } x \in (x_{lb}, x_{ub})
\end{align}
where $(x_{lb}, x_{ub})$ are bounds chosen to reflect realistic values for the problem. This ensures we only lift transitions to the discrete model if they are possible. After integration, we are left with a discrete MDP which contains averaged information about all of the underlying continuous quantities. This includes information about the transitions that the structure of the task space allows, and their corresponding approximate control costs (see \ref{app:hier-decomp}). Note that after each batch update of the rSLDS parameters, this discrete planner must be refitted accordingly.
 
The lifted discrete generative model has all the components of a standard POMDP in the active inference framework:
\begin{equation}
    P(o_{1:T}, s_{1:T}, A, B, \pi) = P(\pi)P(A)P(B)P(s_0)\prod_t P(s_t \mid s_{t-1}, B, \pi)P(o_t \mid s_t, A)
\end{equation}
along with prior over policies $P(\pi)=Cat(E)$, and preference distribution $\tilde P(o_t)=Cat(C)$. Specifically our lifted $P(\pi)$ reflects the approximate control costs of each continuous transition and $\tilde P(o_t)$ reflects the reward available in each mode. We assume an identity mapping between states and observation meaning the state information gain term in Eq.~\ref{G_decomposed_fully} collapses into a maximum entropy regulariser, while we maintain Dirichlet priors over the transition parameters $B$, facilitating directed exploration. Due to conjugate structure Bayesian updates amount to a simple count-based update of the Dirichlet parameters \cite{murphy2012machine}. At each time step, the discrete planner selects a policy by sampling from the following distribution:
\begin{align}
    Q(\pi) &= softmax(-G(\pi) + \ln P(\pi) ).
\end{align}
The policy is then communicated to the continuous controller. Specifically, the first action of the selected policy is a requested transition $i \rightarrow j$ and is translated into a continuous control prior $\tilde p(x)\sim N(x_j, \Sigma_j)$ via the following link function,
\begin{align}
\label{gen-ctl-prior-eq}
x_j &= \underset{x}{\arg\max} \, P(z=j \mid x, u)
\end{align}

whereby we numerically optimise for a point in space up to some probability threshold, $T$ (for details on this optimisation, see \ref{control-priors}). These priors represents an approximately central point in the desired discrete region $j$ requested by the action $a^j$. Note that these priors only need to be calculated once per refit of the rSLDS. The discrete planner infers its current state $s_{\tau}$ from observing $z_t$. Importantly, the discrete planner is only triggered when the system switches into a new mode\footnote{Or a maximum dwell-time (hyperparameter) is reached.}. In this sense, discrete actions are temporally abstracted and decoupled from continuous clock-time in a method reminiscent of the options framework \cite{SUTTON1999181}.

\subsection{Continuous controller}
\label{continuous-controller}
Continuous closed-loop control is handled by a set of continuous active inference controllers. For controlling the transition from mode $i$ to mode $j$ ($x_i$ to $x_j$), the objective of the controller is to minimise the following (discrete-time) expected free energy functional \footnote{As shown in \cite{koudahlEpistemicsExpectedFree2021a} linear state space models preclude state information gain terms leaving the simplified form seen here.}:
\begin{equation}
    G_{ij}(\pi)= \mathbb{E}_{q(\cdot \mid x_0=x_i, \pi)}[(x_S - x_j)^TQ_f(x_S-x_j) + \sum_{t=0}^{S} u_t^T (R - \Pi_t^u) u_t ] + \ln \det \Pi
\end{equation}
Where $S$ is the finite time horizon and the quadratic terms derive from Gaussian preferences about the final state $\tilde{p}_j(x_S) \sim N(x_j, Q_f^{-1})$ and time invariant control input prior $p(u_t) \sim N(0, R^{-1})$ (\ref{app:cts-aif}). Importantly we design the control priors such that the controller only provides solutions within the environments given constraints (for further discussion, see Sec.~\ref{discussion}). The approximate closed-loop solution to each of these sub-problems is computed offline each time the rSLDS is refitted (see \ref{LQR}) using the updated parameters of the linear dynamical systems, allowing for fast discrete-only planning when online.

\section{Results}
\label{methods}

\begin{figure}[b]
    \vskip 0.2in
    \begin{center}
        \centerline{\includegraphics[width=0.8\textwidth]{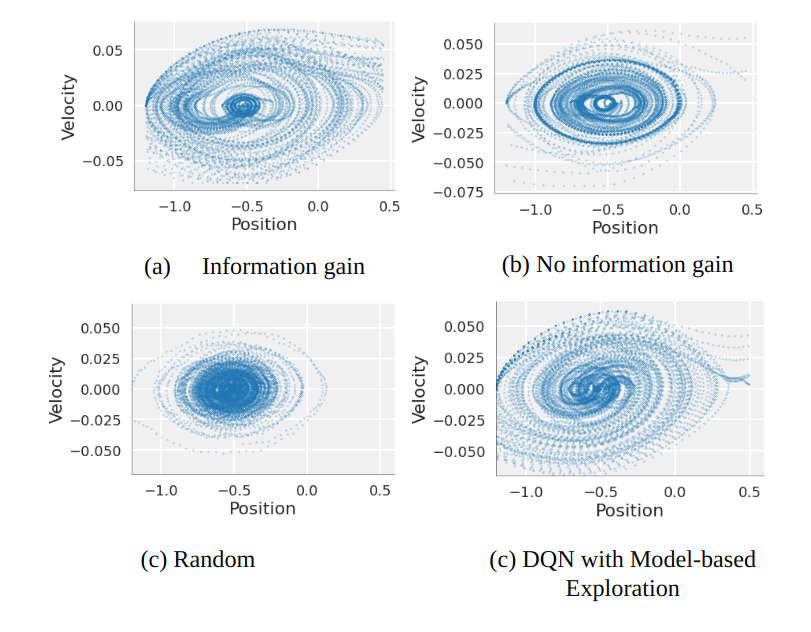}}
        \caption{\textbf{HHA with information-gain explored a wider range of the state-space}. State-space coverage in Continuous Mountain Car after 10,000 steps and best of 3 runs for (a) HHA with information-gain drive, (b) HHA without information gain drive and (c) randomly sampled continuous actions baseline. HHA with information-gain drive also shows comparable performance to (d) a Deep Q-Network with Model-Based Exploration (DQN-MBE) on the (comparably easier) Discrete Mountain Car task \cite{Gou2019}. Exact parameters for DQN-MBE are given in Table~\ref{dqn-table} in \ref{sec-DQNMBE}.}
        \label{fig:exploration}
    \end{center}
    \vskip -0.2in
\end{figure}

To evaluate the performance of our (HHA) model, we applied it to the classic control problem of Continuous Mountain Car. This problem is particularly relevant for our purposes due to the sparse nature of the rewards, necessitating effective exploration strategies to achieve good performance. We find that the HHA finds piecewise affine approximations of the task-space and uses these discrete modes effectively to solve the task. Fig.~\ref{piecewise} shows that while the rSLDS has divided up the space according to position, velocity and control input, the useful modes for solving the task are those found in the position space. Once the goal and a good approximation to the system has been found, the HHA successfully and consistently navigates to the reward. This can be seen in the example trajectories (in Fig.\ref{piecewise}b) where the agent starts at the central position $[0,0]$ and proceeds to rock back and forth within the valley until enough momentum is gained for the car to reach the flag position at a position of 0.5. The episode terminates once the reward has been reached and the agent is re-spawned at the origin before repeating the same successful solution.

Fig.~\ref{fig:exploration} shows that the HHA performs a comprehensive exploration of the state-space and significant gains in the state-space coverage are observed when using information-gain drive in policy selection compared to without. Indeed, our model competes with the state-space coverage achieved by model-based algorithms with exploratory enhancements in the discrete Mountain Car task, which is inherently easier to solve.

\begin{figure}[h!]
    \centering
    \includegraphics[width=0.4\linewidth]{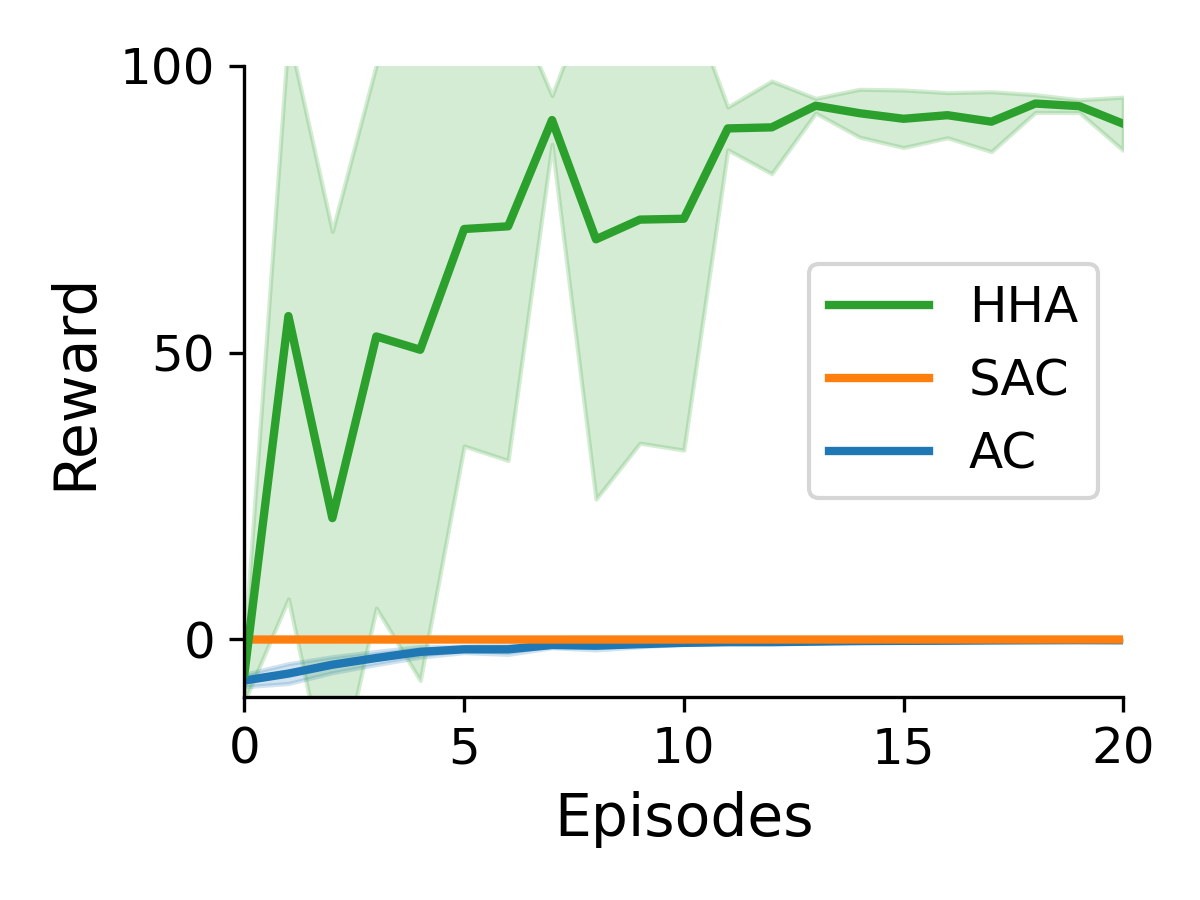}
    \caption{\textbf{HHA both finds the reward and captilises on its experience significantly quicker than model-free RL baselines}. Average reward (+/- std) over 6 runs for Continuous Mountain Car (20 episodes, max episode length of 200 steps) for HHA (our model), Soft-Actor Critic (with 2 Q-functions), and Actor-Critic models. Note that after 20 episodes, SAC and AC are yet to find the reward and converge on a solution.} \label{fig:rewards}
\end{figure}

We compare the performance of the HHA to model-free reinforcement learning baselines (Actor-Critic and Soft Actor-Critic) and find that the HHA both finds the reward and capitalises on its experience significantly quicker than the other models (see Fig.~\ref{fig:rewards}). Given both the sparse nature of the task and the poor exploratory performance of random action in the continuous space, these RL baselines struggle to find the goal within 20 episodes without the implementation of reward-shaping techniques. With reference to the high sample complexity of these algorithms, our model significantly outperforms other baselines in this task.

\section{Discussion}
\label{discussion}

The emergence of non-grid discretisations of the state-space allows us to perform fast systems identification via enhanced exploration, and successful non-trivial planning through the delineation of abstract sub-goals. Hence, the time spent exploring each region is not based on euclidean volume which helps mitigate the curse of dimensionality that other grid-based methods suffer from. Interestingly, even without information-gain, the area covered by our hybrid hierarchical agent is still notably better than that of the random continuous action control (see Fig.~\ref{fig:exploration}c). This is because the agent is still operating at the level of the non-grid discretisation of the state-space which acts to significantly reduce the dimensionality of the search space in a behaviourally relevant way.  

Such a piecewise affine approximation of the space will incur some loss of optimality in the long run when pitted against black-box approximators. This is due to the nature of caching only approximate closed-loop solutions to control within each piecewise region, whilst the discrete planner implements open-loop control. However, this approach eases the online computational burden for flexible re-planning. Hence, in the presence of noise or perturbations within a region, the controller may adapt without any new computation. This is in contrast to other nonlinear model-based algorithms like  model-predictive control where reacting to disturbances requires expensive trajectory optimisation at every step \cite{Schwenzer2021}. By using the piecewise affine framework, we maintain functional simplicity and interpretability through structured representation. We therefore suggest that this method is amenable to future alignment with a control-theoretic approach to safety guarantees for ensuring robust system performance and reliability. Indeed, such use of discrete approximations to continuous trajectories has been shown to improve the ability to handle uncertainty. Evidence of the efficacy of this kind of approach in machine learning applications has been exhibited in recent work by \cite{block2023provableguaranteesgenerativebehavior}, which examined the problem of compounding error in imitation learning from expert demonstration. The authors demonstrated that applying a set of primitive controllers to discrete approximations of the expert trajectory effectively mitigated the accumulation of error by ensuring local stability within each chunk.

We acknowledge there may be better solutions to dealing with control input constraints than the one given in Sec.~\ref{continuous-controller}. Different approaches have been taken to the problem of implementing constrained-LQR control, such as further piecewise approximation based on defining reachability regions for the controller \cite{BEMPORAD20023}.

\section{Conclusion}
In summary, the successful application of our hybrid hierarchical active inference agent in the Continuous Mountain Car problem showcases the potential of recurrent switching linear dynamical systems (rSLDS) for enhancing decision-making and control in complex environments. By leveraging rSLDS to discover meaningful coarse-grained representations of continuous dynamics, our approach facilitates efficient system identification and the formulation of abstract sub-goals that drive effective planning. This method reveals a promising pathway for the end-to-end learning of hierarchical mixed generative models for active inference, providing a framework for tackling a broad range of decision-making tasks that require the integration of discrete and continuous variables. The success of our agent in this control task demonstrates the value of such hybrid models in achieving both computational efficiency and flexibility in dynamic, high-dimensional settings.

\subsubsection*{Acknowledgements}
This work was supported by The Leverhulme Trust through the be.AI Doctoral Scholarship Programme in Biomimetic Embodied AI. Additionally, this research received funding from the European Innovation Council via the UKRI Horizon Europe Guarantee scheme as part of the MetaTool project. We gratefully acknowledge both funding sources for their support.

\subsubsection*{Disclosure of Interests} 
The authors have no competing interests to declare that are relevant to the content of this article.

\newpage

\appendix
\newpage

\section{Appendix}
\label{appendix}

\subsection{Framework}
\label{app:framework}

\textbf{Optimal Control} To motivate our approximate hierarchical decomposition, we adopt the optimal control framework, 
specifically we consider discrete time state space dynamics of the form:
\begin{equation}
    x_{t+1} = f(x_t, u_t, \eta_t)
\end{equation}
with known initial condition $x_0$, and noise $\eta_t$ drawn from some time invariant distribution $\eta_t \sim D$, where we assume $f$ to be $p(x_{t+1} \mid x_t, u_t)$ and is a valid probability density throughout. We use  $c_t: X \times U \rightarrow  \mathbb{R}$ for the control cost function at time $t$ and let $\mathbb{U}$ be the set of admissible (non-anticipative, continuous) feedback control laws, possibly restricted by affine constraints. The optimal control law for the finite horizon problem is given as: 
\begin{align}
    J(\pi) &= \mathbb{E}_{x_{0},\pi}[\sum_{t=0}^{T} c_t(x_t, u_t)] \\
    \pi^{*} &= \arg \min_{\pi \in \mathbb{U}} J(\pi) 
\end{align}

\textbf{PWA Optimal Control} The fact we do not have access to the true dynamical system $f$ motivates the use of a piecewise affine (PWA) approximation. Also known as hybrid systems:
\begin{align}
    x_{t+1} &= A_{i} x_t + B_{i} u_t + \epsilon_t \\
    & \text{when } (x_t, u_t) \in H_i
\end{align}
Where $\mathbb{H}=\{H_i: i \in [K] \}$ is a polyhedral partition of the space $X\times U$. In the case of a quadratic cost function, it can be shown the optimal control law for such a system is peicewise linear. Further there exist many completeness (universal approximation) type theorems for peicewise linear approximations implying if the original system is controllable, there will exist a peicewise affine approximation through which the system is still controllable \cite{Bemporad2000,Borrelli2006}.

\textbf{Relationship to rSLDS} We perform a canonical decomposition of the control objective $J$ in terms of the components or modes of the system. By slight abuse of notation $[x_t = i]:=[(x_t, u_t) \in H_i]$ represent the Iverson bracket.
\begin{align}
    J(\pi) &=  \sum_t \int p_{\pi}(x_t \mid x_{t-1}, u_t)c_t(x_t, u_t) d x_{t} dx_{t-1} \\
    &= \sum_t \int \sum_{i\in [K]} [x_{t-1} = i]p_{\pi}(x_t \mid x_{t-1}, u_t)c_t(x_t, u_t) d x_{t} dx_{t-1}
\end{align}
Now let $z_t$ be the random variable on $[K]$ induced by $Z_t = i$ if $[x_t = i]$ we can rewrite the above more concisely as,
\begin{align} 
    J(\pi) &=\sum_t \int \sum_{i\in [K]} p_{\pi}(x_t, z_{t-1}=i \mid x_{t-1}, u_t)c_t(x_t, u_t) d x_{t} dx_{t-1} \\
    &= \sum_{i\in [K]}\sum_t \int  p_{\pi}(x_t, z_{t-1}=i \mid x_{t-1}, u_t)c_t(x_t, u_t) d x_{t} dx_{t-1} \\
    &=\sum_{i\in [K]}\sum_t \mathbb{E}_{\pi_i}[c_t(x_t, u_t)] \label{eq:decomp}
\end{align}
which is just the expectation under a recurrent dynamical system with deterministic switches. Later (see \ref{online-problem}), we exploit the non-deterministic switches of rSLDS in order to drive exploration. Eq.\ref{eq:decomp} demonstrates the global problem can be partitioned solving problems within each region (inner expectation), and a global discrete problem which decides which sequence of regions to visit. In the next section, we introduce a new set of variables which allows us to approximately decouple the problems.

\subsection{Hierarchical Decomposition}
\label{app:hier-decomp}

Our aim was to decouple the discrete planning problem from the fast low-level controller. In order to break down the control objective in this manner, we first create a new discrete variable which simply tracks the transitions of $z$, this allows the discrete planner to function in a temporally abstracted manner.

\textbf{Decoupling from clock time} Let the random variable $(\zeta_s)_{s>0}$ record the transitions of $(z_t)_{t>0}$ i.e. let 
\begin{equation}
    \tau_s(\tau_{s-1})= \min \{t: z_{t+1} \neq z_t,  t> \tau_{s-1}\}, \tau_0=0
\end{equation}
be the sequence of first exit times, then $\zeta$ is given by $\zeta_s = z_{\tau_s}$. With these variables in hand, we frame a small section of the global problem as a first exit problem.

\textbf{Low level problems} Consider the first exit problem for exiting region $i$ and entering $j$ defined by: 
\begin{align}
    \pi_{ij}(x_0) &= \arg \min_{\pi, S} J_{ij}(\pi, x_0, S) \\
    J_{ij}(\pi, x_0, S)&= \mathbb{E}_{\pi, x_0}[\sum_{t=0}^{S}c(x_t, u_t)]\\
    & \text{s.t. } (x_t, u_t) \in H_i \\
    & \text{s.t. } c(x, u) = 0 \text{ when } (x,u) \in \partial H_{ij}
\end{align}
where $\partial H_{ij}$ is the boundary $H_i \bigcap H_j$. Due, to convexity of the polyhedral partition, the full objective admits the decomposition in terms of these subproblems,
\begin{align}
    J(\pi) &= \sum_s J_{\zeta(s+1), \zeta(s)}(\pi, x_{t_s}, t_{s+1} - t_s)
\end{align}
Ideally, we would like to simply solve all possible subproblems $\{J_{ij}^*(x): i,j \in [K]\times [K]\}$ and then find a sequence of discrete states,  $\zeta(1),...,\zeta(S)$, which minimises the sum of the sub-costs, however notice each sub-cost depends on the starting state, and further this is determined by the final state of the previous problem. A pure separation into discrete and continuous problems is not possible without a simplifying assumption.  

\textbf{Slow and fast mode assumption} The goal is to tackle the decomposed objectives individually, however the hidden constraint that the trajectories line up presents a computational challenge. Here we make the assumption that the difference in cost induced by different starting positions within a region is much less than expected difference in cost  of starting in a different region. This assumption justifies using an average cost for the low-level problems to create the high-level problem.

\textbf{High level problem} we let $J_{ij}^* = \min_{\pi} \int_{x_0}J_{ij}(\pi, x_0)p(x_0)$ be the average cost of each low-level problem. We form a Markov decision process by introducing abstract actions $a \in [K]$:
\begin{equation}
    p_{ik}(a) = \mathbb{P}(\zeta_{s+1}=k \mid \zeta_{s}=i, a=j, \pi_{ij}^*)
\end{equation}
and let $p_{\pi_d}$ be the associated distribution over trajectories induced by some discrete state feedback policy, along with the discrete state action cost $c_d(a=j, \zeta=i) = J_{ij}^*$ we may write the high level problem:
\begin{align}
    \pi_d^* &= \min_{\pi_d} J_d(\pi, \zeta_0) \\
    J_d(\pi, \zeta_0)&=\mathbb{E}p_{\pi_d, \zeta_0}[\sum_{s=0}^{S} c_d( a_s, \zeta_s)]
\end{align}
Our overall approximate control law is then given by choosing the action of the continuous controller $\pi_{ij}(x)$ suggested by the discrete policy $\pi_d(i(x))$, or more concisely,  $\pi(x) = \pi_{i(x), \pi_d^* \circ i(x)}(x)$, where $i$ is calculates the discrete label (MAP estimate) for the continuous state $x$. In the next sections we describe the methods used to solve the high and low level problems.

\subsection{Offline Low Level Problems: Linear Quadratic Regulator (LQR)}
\label{LQR}
Rather than solve the first-exit problem directly, we formulate an approximate problem by finding trajectories that end at specific `control priors' (see \ref{control-priors}).
Recall the low level problem given by:
\begin{align}
    \pi_{ij}(x_0) &= \arg \min_{\pi, S} J_{ij}(\pi, x_0, S) \\
    J_{ij}(\pi, x_0, S)&= \mathbb{E}_{\pi, x_0}[\sum_{t=0}^{S}c(x_t, u_t)]\\
    & \text{s.t. } (x_t, u_t) \in H_i \\
    & \text{s.t. } c(x, u) = 0 \text{ when } (x,u) \in \partial H_{ij}
\end{align}
In order to approximate this problem with one solvable by a finite horizon LQR controller, we adopt a fixed goal state, $x^* \in H_j$. Imposing costs $c_t(x_t, u_t) = u_t^T R u_t$ and $c_S(x_S, u_S) = (x - x^*) Q_f (x - x^*)$. Formally we solve,
\begin{align}
    \pi_{ij}(x_0) &= \arg \min_{\pi, S} J_{ij}(\pi, x_0, S) \\
    J_{ij}(\pi, x_0, S)&= \mathbb{E}_{\pi, x_0}[(x_S - x^*)^T Q_f (x_S - x^*) + \sum_{t=0}^{S-1} u_t^T R u_t]\\
\end{align}
by integrating the discrete Ricatti equation backwards. Numerically, we found optimising over different time horizons made little difference to the solution, so we opted to instead specify a fixed horizon (hyperparameter). These solutions are recomputed offline every time the linear system matrices change.

\textbf{Designing the cost matrices}
Instead of imposing the state constraints explicitly, we record a high cost which informs the discrete controller to avoid them. In order to approximate the constrained input we choose a suitably large control cost $R=rI$. We adopted this approach for the sake of simplicity, potentially accepting a good deal of sub-optimality. However, we believe more involved methods for solving input constrained LQR could be used in future, e.g. \cite{Bemporad2000}, especially because we compute these solutions offline.

\subsection{Active Inference Interpretation} \label{app:cts-aif}
\subsubsection{Expected Free Energy} 
Here we express the fully-observed continuous (discrete time) active inference controller, without mean-field assumptions, and show it reduces to a continuous quadratic regulator.
Suppose we have a linear state space model:
\begin{equation}
    x_{t+1} = A x_{t} + B u_{t} + \epsilon_t
\end{equation}
and a prior preference over trajectories $\tilde p(x_{1:T}) \sim N(x_T; x_f, Q_f^{-1})$, active inference specifies the agent minimises
\begin{equation}
    G(\pi)= \mathbb{E}_{q(x_{1:T}, u_{1:T}; \pi)}[-\ln \tilde{p}(x_{1:T}, u_{1:T})+ \ln q(x_{1:T}, u_{1:T}; \pi)]
\end{equation}
Note, since all states are fully observed we have no ambiguity term.
Where $\tilde{p}(x_{1:T}, u_{1:T}) \propto \tilde{p}(x_{1:T})p(x_{1:T} \mid u_{1:T})p(u_{1:T})$,  the central term is the dynamics model and the prior over controls is also gaussian, $p(u_{1:T}) = \prod_t N(u_t; 0, R^{-1})$.
Finally, we adopt $q(x_{1:T}, u_{1:T}; \pi) = p(x_{1:T} \mid u_{1:T})\prod_t \pi_t(u_t \mid x_t)$, where we parametrise the variational distributions as $\pi_t \sim N(u_t; K_tx, \Sigma_t^q)$ (where $K_t, \Sigma_t^)$ are parameters to be optimised).
The expected free energy thus simplifies to:
\begin{equation}
    G(\pi)= \mathbb{E}_{q(x_{1:T}, u_{1:T}; \pi)}[(x_T - x_F)^TQ_f(x_T-x_F) - \sum_t u_t^T (R + \Pi_t^u) u ] + \ln \det \Pi
\end{equation}
\subsubsection{Dynamic Programming (HJB)}
We proceed by dynamic programming, let the `value' function be  
\begin{equation}
    V(x_{k}) = \min_{ \pi } \mathbb{E}_{q(x_{k+1:T}, u_{k:T} \mid x_k, \pi)}[(x_T - x_f)^TQ_f(x_T - x_f) + \sum_{t=k}^T u_t^T (R + \Pi_t^u) u_t] + \ln \det \Pi
\end{equation}
As usual the value function satisfies a recursive property:
\begin{equation}
    V(x_{k}) = \min_{ \pi } \mathbb{E}_{q(x_{k+1}, u_{k} \mid x_k, \pi)}[u_k^T (R + \Pi_k^u) u_k + V(x_{k+1})]  + \ln \det \Pi
\end{equation}
We introduce the ansatz $V(x_{k}) = x_k^T S_k x_k$ leading to,
\begin{equation} \label{eqn:recursion}
    x_{k}^TS_k x_{k} = \min_{ \pi } \mathbb{E}_{q(x_{k+1}, u_{k} \mid x_k, \pi)}[u_k^T (R + \Pi_k^u) u_k + x_{k+1}^TS_{k+1} x_{k+1}]  + \ln \det \Pi
\end{equation}
Finally we take expectations, which are available in closed form, and solve for $\Sigma_k$ and $K_k$:
\begin{align}
    x_{k}^TS_{k}x_{k} = \min_{K_k, \Pi_k} & x_k^TK_k^T R K_k x_k + tr(\Sigma_k^u (R + \Pi_t^u)) \\
    &+ x_k^T(A+BK_k)^T S_{k+1}(A+BK_k)x_k + tr(\Sigma_k^x S_{k+1})  + \ln \det \Pi
\end{align}
Solving for $\Sigma_k$ and substituting,
\begin{equation}
    \Sigma_k^q = (R+\Pi_k^u)^{-1}
\end{equation}
\begin{equation}
   \implies S_{k} = \min_{K_k} K_k^T R K_k  + (A+ BK_k)^T S_{k+1}(A+ BK_k) 
\end{equation}
\begin{equation}
    K_k = - (R + B^T S_{K+1} B)^{-1}B^TS_{K+1} A
\end{equation}
Where $S_k$ follows the discrete algebraic Riccati equation (DARE). 

Thus we recover $\pi_t(u \mid x) \sim N(K_tx, \Sigma_k)$ where $K_t$ is the traditional LQR gain, and $\Sigma_t$ solves $\Sigma_k = (R+\Pi_k)^{-1}$. Here we use the deterministic maximum-a-posterori `MAP' controller $K_tx$. However the collection of posterior variance estimates adds a different total cost depending on the variance inherent in the dynamics which can be lifted to the discrete controller.

\subsubsection{As Belief Propagation}
A different perspective is as message passing:
we wish to calculate the marginals $p(x_k)$ and $p(x_k, u_k)$ tilted by the preference distribution $\tilde{p}(x_k)$ and control prior $p(u)$ for this we can integrate backwards using the recursive formula
\begin{align}
    b(x_k) &= \int b(x_k, u_k) d x_k \\ 
    b(x_k, u_k) &= \int \tilde{p}(x_k) p(x_{k+1} \mid x_k, u_k )p(u_k) b(x_{k+1}) d x_{k+1}
\end{align}
from which we can extract the control law $p(u_k \mid x_k) = b(x_k, u_k)/b(x_k)$. To proceed we use the variational method to marginalise:
\begin{equation}
    - \ln b(x_k) =  \min_{q} \mathbb{E}_{q} [-\ln \tilde{p}(x_k, u_k)p(x_{k+1} \mid x_k, u_k )b(x_{k+1}) + \ln q(x_{k+1}, u_k \mid x_k)]
\end{equation}
making the same assumption as above about variational distributions, and introducing the ansatz $b(x_k) \sim N(x_k; 0, S_k)$ leads to the same equation as \ref{eqn:recursion} up to irrelevant constants.

\subsection{Online high level problem}
\label{online-problem}
The high level problem is a discrete MDP with a `known' model, so the usual RL techniques (approximate dynamic programming, policy iteration) apply. Here, however we choose to use a model-based algorithm with a receding horizon inspired by Active Inference, allowing us to easily incorporate exploration bonuses.

Let the Bayesian MDP be given by $\mathcal{M}_B = (S, A, P_a, R, P_\theta)$ be the MDP, where $p_a(s_{t+1}\mid s_t, a_t, \theta)\sim Cat(\theta_{as})$ and $p(\theta_{as}) \sim Dir(\alpha)$. We estimate the open-loop reward plus optimistic information-theoretic exploration bonuses.

\textbf{Active Inference conversion}
We adopt the Active Inference framework for dealing with exploration. Accordingly we adopt the notation $\ln \tilde{p}(s_{t}, a_t) = R(s_t, a_t)$ and refer to this 
`distribution' as the goal prior \cite{millidge2020relationship}, and optimise over open loop policies $\pi = (a_0, ..., a_T)$.

\begin{equation}
    G(a_{1:T}, s_0) = \mathbb{E}[\sum_{t=0}^{T} R(s_t, a_t) + IG_{p} + IG_{s}  \mid s_0, {a_{1:T}}]
\end{equation}
where parameter information-gain is given by $IG_{p} = D_{KL}[p_{t+1}(\theta) \mid \mid p_t(\theta)]$, with $p_t(\theta) = p(\theta \mid s_{0:t})$. In other words, we add a bonus when we expect the posterior to diverge from the prior, which is exactly the transitions we have observed least \cite{heins2022pymdp}.

We also have a state information-gain term, $IG_s = D_{KL}[p_{t+1}(s_{t+1}) \mid \mid p_t(s_{t+1})]$. In this case (fully observed), $p_{t+1}(s_{t+1}) = \delta_s$ is a one-hot vector. Leaving the term $\mathbb{E}_{t}[-\ln p_t(s_{t+1})]$ leading to a maximum entropy term \cite{heins2022pymdp}.

We calculate the above with Monte Carlo sampling which is possible due to the relatively small number of modes. Local approximations such as Monte Carlo Tree Search could easily be integrated in order to scale up to more realistic problems. Alternatively, for relatively stationary environments we could instead adopt approximate dynamic programming methods for more habitual actions. 

\subsection{Generating continuous control priors}
\label{control-priors}
In order to generate control priors for the LQR controller which correspond to each of the discrete states we must find a continuous state $x_i$ which maximises the probability of being in a desired $z$:

\begin{align}
x_i = \underset{x}{\arg\max} \, P(z=i|x, u)
\end{align}

For this we perform a numerical optimisation in order to maximise this probability. Consider that this probability distribution $P(z = i |x)$  is a softmax function for the i-th class is defined as:

\begin{align}
\sigma(v_i) = \frac{\exp (v_i)}{\sum_j \exp (v_j)}, v_i = w_i \cdot x + r_i
\end{align}

where $w_i$ is the i-th row of the weight matrix, $x$ is the input and $r_i$ is the i-th bias term. The update function used in the gradient descent optimisation can be described as follows:

\begin{align}
x \leftarrow x + \eta \nabla_x \sigma(v_i)
\end{align}

where $\eta$ is the learning rate and the gradient of the softmax function with respect to the input vector $x$ is given by:

\begin{align}
\nabla_x \sigma(v_i) = \frac{\partial \sigma(v_i)}{\partial v} \cdot \frac{\partial v}{\partial x} = \sigma(v_i)(\bold e_i - \sigma(v))\cdot W
\end{align}

in which $\sigma(v)$ is the vector of softmax probabilities, and $\bold e_i$ is the standard basis vector with 1 in the i-th position and 0 elsewhere. The gradient descent process continues until the probability $P(z=i | x)$ exceeds a specified threshold $\theta$ which we set to be 0.7. This threshold enforces a stopping criterion which is required for the cases in which the region $z$ is unbounded.

\subsection{Model-free RL baselines}

\begin{table}[H]
\caption{Summary of the Soft Actor-Critic algorithm with multiple Q-functions.}
\label{sample-table-sac}
\vskip 0.15in
\begin{center}
\begin{small}
\begin{sc}
\begin{tabular}{lcccr}
\toprule
Component & Input \\
\midrule
Q-network    & 3×256×256×256×2\\
Policy network & 2×256×256×256×2\\
Entropy regularization coeff    & 0.2\\
Learning rates (Qnet + Polnet)    & 3e-4\\
Batchsize     & 60\\
\bottomrule
\end{tabular}
\end{sc}
\end{small}
\end{center}
\vskip -0.1in
\end{table}

\begin{table}[H]
\caption{Summary of the Actor-Critic algorithm}
\label{sample-table-ac}
\vskip 0.15in
\begin{center}
\begin{small}
\begin{sc}
\begin{tabular}{lcccr}
\toprule
Component & Input \\
\midrule
Feature Processing    & StandardScaler, RBF Kernels (4 $\times$ 100) \\
Value-network    & 4001 parameters (1 dense layer)\\
Policy network & 802 parameters (2 dense layers) \\
Gamma & 0.95\\
Lambda    & 1e-5\\
Learning rates (Policy + Value)    & 0.01\\
\bottomrule
\end{tabular}
\end{sc}
\end{small}
\end{center}
\vskip -0.1in
\end{table}

\subsection{Model-based RL baseline}
\label{sec-DQNMBE}

\begin{table}[H]
\caption{Summary of DQN-MBE algorithm \cite{Gou2019}}
\label{dqn-table}
\vskip 0.15in
\begin{center}
\begin{small}
\begin{sc}
\begin{tabular}{lcccr}
\toprule
Component & Input \\
\midrule
Q-network    & 1 hidden-layer, 48 units, ReLU \\
Dynamics Predictor Network (Fully Connected) & 2 hidden-layers (each 24 Units), ReLU \\
$\epsilon$ minimum    & 0.01 \\
$\epsilon$ decay    & 0.9995\\
Reward discount & 0.99 \\
Learning rates (Qnet / Dynamics-net)    & 0.05 / 0.02\\
Target Q-network update interval & 8\\
Initial exploration only steps     & 10000\\
Minibatch size (Q-network)   & 16\\
Minibatch size (dynamics predictor network)     & 64\\
Number of recent states to fit probability model     & 50\\
\bottomrule
\end{tabular}
\end{sc}
\end{small}
\end{center}
\vskip -0.1in
\end{table}

\newpage
%
\bibliographystyle{splncs04}
\bibliography{thebibliography}

\end{document}